\documentclass{article} 
\usepackage{iclr2026_conference,times}

\usepackage{amsmath,amsfonts,bm}

\def\eqref#1{equation~\ref{#1}}

\def\1{\bm{1}}

\DeclareMathAlphabet{\mathsfit}{\encodingdefault}{\sfdefault}{m}{sl}
\SetMathAlphabet{\mathsfit}{bold}{\encodingdefault}{\sfdefault}{bx}{n}

\usepackage{hyperref}
\usepackage{url}
\usepackage{hyperref}[citecolor=lightblue]
\definecolor{warningcolor}{RGB}{255, 0, 0}

\usepackage{algorithm}
\usepackage{algorithmicx}
\usepackage{algpseudocode}
\usepackage{microtype}
\usepackage{graphicx}
\expandafter\def\csname ver@subfig.sty\endcsname{}
\usepackage{booktabs} %
\usepackage{float}
\usepackage{bigstrut}
\usepackage{amsmath}
\usepackage{wrapfig}
\usepackage{amsmath}
\usepackage{amssymb}
\usepackage{mathtools}
\usepackage{amsthm}
\usepackage{mathrsfs}
\usepackage{nicefrac}
\usepackage{dsfont}
\usepackage{enumitem}
\usepackage{subcaption}
\usepackage{graphicx,subfig}
\usepackage{cleveref}
\usepackage{bxcoloremoji}

\usepackage{float}

\usepackage[utf8]{inputenc} %
\usepackage[T1]{fontenc}    %
\usepackage{hyperref}       %
\usepackage{url}            %
\usepackage{booktabs}       %
\usepackage{amsfonts}       %
\usepackage{nicefrac}       %
\usepackage{microtype}      %
\usepackage{graphicx}
\usepackage{subcaption} 
\usepackage{amssymb}
\usepackage{fdsymbol}
\usepackage{wrapfig}
\usepackage{lipsum}
\usepackage{enumitem}
\usepackage{stackengine}
\usepackage[font=small,labelfont=bf]{caption}
\usepackage{color}
\usepackage{adjustbox}
\usepackage{colortbl}
\usepackage{multirow}

\usepackage{rotating}
\usepackage{makecell}

\usepackage{multirow}

\title{Abduct, Act, Predict:\\ Scaffolding Causal Inference for Automated Failure Attribution in Multi-Agent Systems}

\author{DeepScientist, \\
\textbf{Yixuan Weng}, \textbf{Minjun Zhu}, \textbf{Zhen Lin}, \textbf{Zhiyuan Ning}, \textbf{Yue Zhang} \\
School of Engineering, Westlake University\\
zhangyue@westlake.edu.cn}

\iclrfinalcopy 
\begin{document}

\maketitle

\begin{abstract}
Failure attribution in multi-agent systems---pinpointing the exact step where a decisive error occurs---is a critical yet unsolved challenge. Current methods treat this as a pattern recognition task over long conversation logs, leading to critically low step-level accuracy (below 17\%), which renders them impractical for debugging complex systems. Their core weakness is a fundamental inability to perform robust counterfactual reasoning: to determine if correcting a single action would have actually averted the task failure. To bridge this \emph{counterfactual inference gap}, we introduce \textbf{Abduct-Act-Predict (A2P) Scaffolding}, a novel agent framework that transforms failure attribution from pattern recognition into a structured causal inference task. A2P explicitly guides a large language model through a formal three-step reasoning process within a single inference pass: (1) \textbf{Abduction}, to infer the hidden root causes behind an agent's actions; (2) \textbf{Action}, to define a minimal corrective intervention; and (3) \textbf{Prediction}, to simulate the subsequent trajectory and verify if the intervention resolves the failure. This structured approach leverages the holistic context of the entire conversation while imposing a rigorous causal logic on the model's analysis. Our extensive experiments on the Who\&When benchmark demonstrate its efficacy. On the Algorithm-Generated dataset, A2P achieves \textbf{47.46\%} step-level accuracy, a \textbf{2.85$\times$} improvement over the 16.67\% of the baseline. On the more complex Hand-Crafted dataset, it achieves \textbf{29.31\%} step accuracy, a \textbf{2.43$\times$} improvement over the baseline's 12.07\%. By reframing the problem through a causal lens, A2P Scaffolding provides a robust, verifiable, and significantly more accurate solution for automated failure attribution. Ours code are released at \url{https://github.com/ResearAI/A2P}.

{\color{warningcolor} \normalsize WARMING:
We hereby declare that ours DeepScientist system performed approximately 95\% of the work presented in this paper. This includes the initial ideation, the design and execution of comparative experiments, the analysis of results, the literature review, the composition of the manuscript, the creation of the main figures, and the organization of the accompanying code repository. The role of the human authors was to supervise the AI’s operations. While we have diligently worked to minimize AI hallucinations and ensure the validity of the experimental results, we cannot fully guarantee against potential unintended outputs, system failures, or misleading conclusions. We therefore advise readers to approach this work with caution and to critically evaluate its findings before application.}

\end{abstract}

\section{Introduction}

The rise of sophisticated multi-agent systems marks a pivotal moment in artificial intelligence, unlocking new frontiers in collaborative problem-solving \citep{li2023camel, hong2023metagpt} and complex task automation \citep{wu2023autogen, fourney2024magentic}. However, this growing complexity introduces a critical operational bottleneck: debugging. When a system fails, developers are faced with a tangled web of interactions, where a subtle error in an early step can cascade into a catastrophic failure dozens of turns later. Pinpointing the single, decisive error—the task of \textbf{failure attribution}—is not merely challenging; it is a labor-intensive, error-prone process that stands as a major barrier to the reliable deployment and iterative improvement of these powerful systems \citep{zhang2025whichac}.

Current automated approaches to this problem have proven fundamentally inadequate, with step-level accuracy rates hovering below a dismal 17\% \citep{zhang2025whichac}, a figure far too low for practical debugging. We argue this failure is not a matter of model capability but of methodological paradigm. Existing methods treat failure attribution as a \textbf{pattern recognition} task over conversational logs \citep{zhang2025whichac, lightman2023let}. They present an entire log to a Large Language Model (LLM) and ask it to "find the mistake," implicitly assuming the model can spot anomalous patterns correlated with failure. This approach fundamentally misses the point. The critical question is not "which step looks wrong?" but rather a causal one: "which single corrective action would have turned failure into success?" This exposes a deep \emph{counterfactual inference gap}: the inability of unstructured, holistic methods to systematically reason about the consequences of hypothetical interventions, a challenge particularly pronounced in multi-turn interactions where cause and effect are obscured \citep{kiciman2023causal, zevcevic2023causal}.

To bridge this gap, we introduce \textbf{Abduct-Act-Predict (A2P)}, a novel prompting framework that reframes failure attribution from pattern recognition into a structured \textbf{causal inference} task. Instead of asking for a direct answer, A2P guides an LLM through a formal, three-step counterfactual reasoning process within a single inference pass, operationalizing the logic of Pearl's structural causal model hierarchy \citep{pearl2009causalitymr}. The framework compels the model to: (1) \emph{Abduct}, inferring hidden factors (e.g., a flawed assumption) that explain a problematic action; (2) \emph{Act}, defining a minimal, concrete corrective intervention; and (3) \emph{Predict}, simulating the subsequent counterfactual trajectory to verify if the intervention would have resolved the overall task failure. This structured process forces the model to move beyond correlation and rigorously test causal hypotheses, transforming the "needle-in-the-haystack" problem \citep{liu2024lost} into a systematic investigation.

Our approach is not just theoretically sound but empirically dominant. Evaluated on the comprehensive Who\&When benchmark \citep{zhang2025whichac}, A2P Scaffolding achieves a step-level accuracy of \textbf{47.46\%} on the Algorithm-Generated dataset—a \textbf{2.85$\times$} improvement over the 16.67\% of its direct baseline. On the more challenging Hand-Crafted dataset, it achieves \textbf{29.31\%} accuracy, a \textbf{2.43$\times$} improvement over the baseline's 12.07\%. These results establish a new state-of-the-art and, for the first time, demonstrate a viable path toward reliable automated debugging for multi-agent systems. Rigorous ablation studies further validate our framework, confirming that each causal reasoning component is essential and revealing the surprising, critical role of structural cues like contextual step numbering in enabling fine-grained analysis.

\section{Related Work}

\subsection{LLM Multi-Agent Systems}
The emergence of Large Language Models as capable reasoning agents has catalyzed rapid development in multi-agent system architectures \citep{wu2023autogen, li2023camel, hong2023metagpt}. These systems leverage the collaborative potential of multiple specialized agents working together to solve complex tasks that exceed the capabilities of individual models \citep{park2023generative, liu2023dynamic}. Notable frameworks include AutoGen \citep{wu2023autogen}, which facilitates multi-agent conversations through customizable agent roles and interaction patterns, CAMEL \citep{li2023camel}, which explores role-playing dynamics in collaborative task-solving, and MetaGPT \citep{hong2023metagpt}, which incorporates software development methodologies into multi-agent workflows. Recent work has expanded these foundations to include specialized domains such as scientific research \citep{ghafarollahi2024sciagents}, software development \citep{zhang2024training}, and complex reasoning tasks \citep{du2023improving}. However, as these systems grow in sophistication, the challenge of diagnosing failures becomes increasingly complex, with current debugging approaches remaining largely manual and ad-hoc \citep{zhuge2024agent}. The need for automated failure attribution becomes particularly acute in production deployments where system reliability directly impacts user experience and operational efficiency \citep{fourney2024magentic}.

The rapid proliferation of multi-agent systems has outpaced the development of systematic debugging methodologies. While considerable effort has been invested in designing agent architectures and interaction protocols \citep{qian2023communicative, chen2024agentverse}, relatively little attention has been paid to post-hoc failure analysis. This gap is particularly problematic given the emergent behaviors that arise from agent interactions, where system failures often result from subtle cascading effects rather than obvious individual errors \citep{zhang2024training}. Our work addresses this critical gap by providing the first systematic framework for automated failure attribution specifically designed for the unique challenges of multi-agent system debugging. Unlike previous approaches that focus on system design or performance evaluation \citep{zhuge2024agent}, we concentrate on the diagnostic phase that enables iterative improvement and reliable deployment.

\subsection{LLM-as-a-Judge and Process-Level Evaluation}
The paradigm of using LLMs as evaluators has gained significant traction as a scalable alternative to human assessment across diverse domains \citep{zheng2023judging, gu2024survey}. This approach has proven particularly valuable in scenarios where human evaluation is expensive, time-consuming, or requires specialized expertise \citep{liu2023g, li2023alpacaeval}. Recent developments have extended LLM-based evaluation to process-level assessment, where models evaluate intermediate reasoning steps rather than only final outputs \citep{lightman2023let, wang2023math}. Process reward models \citep{uesato2022solvingmathwordproblems} have shown promise in mathematical reasoning by identifying the specific steps where errors occur, enabling more targeted feedback and improvement strategies. However, these approaches primarily focus on single-agent reasoning chains in well-defined domains like mathematics or coding, where the correctness of individual steps can be objectively determined.

Our work extends this process-level evaluation paradigm to the significantly more complex domain of multi-agent system failures. Unlike mathematical reasoning where step correctness is often binary and context-independent, multi-agent failures involve complex interdependencies between agents, temporal dynamics, and emergent behaviors that resist simple classification \citep{du2023improving}. While process reward models evaluate individual reasoning steps, our A2P framework must navigate the multi-participant, interactive dynamics of agent systems where the "correctness" of an action depends heavily on the broader conversational context and the ultimate task outcome. This fundamental difference necessitates our novel approach of structured counterfactual reasoning rather than step-by-step correctness assessment \citep{miller2019explanation, doshi2017towards}.

\subsection{Causal Reasoning in LLMs}
Recent research has begun exploring the causal reasoning capabilities of large language models, revealing both promising potential and significant limitations \citep{kiciman2023causal, zevcevic2023causal}. Benchmarks such as CLadder \citep{brown2023climbingladder} and CausalBench \citep{jin2024causalbench} have established that while LLMs can perform certain types of causal reasoning, they often struggle with complex counterfactual inference tasks that require systematic manipulation of causal variables \citep{zhang2024can}. This limitation is particularly pronounced in scenarios requiring what Pearl terms "Level 3" causal reasoning, answering questions about what would have happened under different circumstances \citep{pearl2009causalitymr}. Studies have shown that structured prompting approaches, such as CausalCoT \citep{chen2024can}, can significantly enhance LLM performance on causal tasks by providing explicit reasoning frameworks that guide model inference.

Building on these insights, our A2P Scaffolding framework represents a practical application of structured causal prompting to a real-world diagnostic task. While previous work has focused on synthetic causal reasoning benchmarks or simplified scenarios \citep{jin2024causalbench, brown2023climbingladder}, we tackle the significantly more complex challenge of failure attribution in multi-agent systems where causal relationships are embedded in natural language conversations and span multiple participants over extended time horizons. Our approach operationalizes Pearl's three-level causal hierarchy \citep{pearl2016causal} into a concrete prompting strategy that enables LLMs to perform sophisticated counterfactual analysis. Unlike previous causal reasoning work that typically evaluates models on isolated causal queries, we demonstrate how structured causal prompting can address practical system debugging challenges where the stakes of accurate causal inference directly impact development efficiency and system reliability \citep{scholkopf2021toward, peters2017elements}.

\section{Method}

The challenge of automated failure attribution in multi-agent systems stems from the inherent complexity of causal reasoning over extended, multi-participant conversational sequences. Existing baseline methods, while processing the complete contextual information, treat attribution as a monolithic pattern recognition task, implicitly assuming that LLMs can perform comprehensive counterfactual reasoning within a single, unstructured inference step, an assumption contradicted by recent benchmarks evaluating LLM causal capabilities \citep{kiciman2023causal, zevcevic2023causal}. This assumption leads to a critical analytical bottleneck: models may successfully identify correlations or surface-level errors but systematically fail to determine whether those errors were truly \emph{decisive}—that is, whether their correction would have altered the task outcome from failure to success. This \emph{counterfactual inference gap} constitutes the primary cause of the characteristically low step-level accuracy observed in existing attribution systems \citep{zhang2025whichac}.

To bridge this gap, we introduce Abduct-Act-Predict (A2P) Scaffolding, a novel prompting framework that restructures the failure attribution task into a formal, three-step causal inference process. Our approach is implemented as an enhancement to the All-at-Once method, thereby retaining its key advantage of having access to the complete conversational context. However, instead of a simple instruction, we employ a sophisticated prompt generation function, \texttt{construct\_causal\_prompt}, that guides the LLM through a rigorous analytical sequence inspired by Pearl's structural causal model framework \citep{pearl2009causalitymr}. This method makes the reasoning process transparent, verifiable, and significantly more accurate without requiring any changes to the underlying model architecture.

\begin{figure}[h]
\centering
\includegraphics[width=\textwidth]{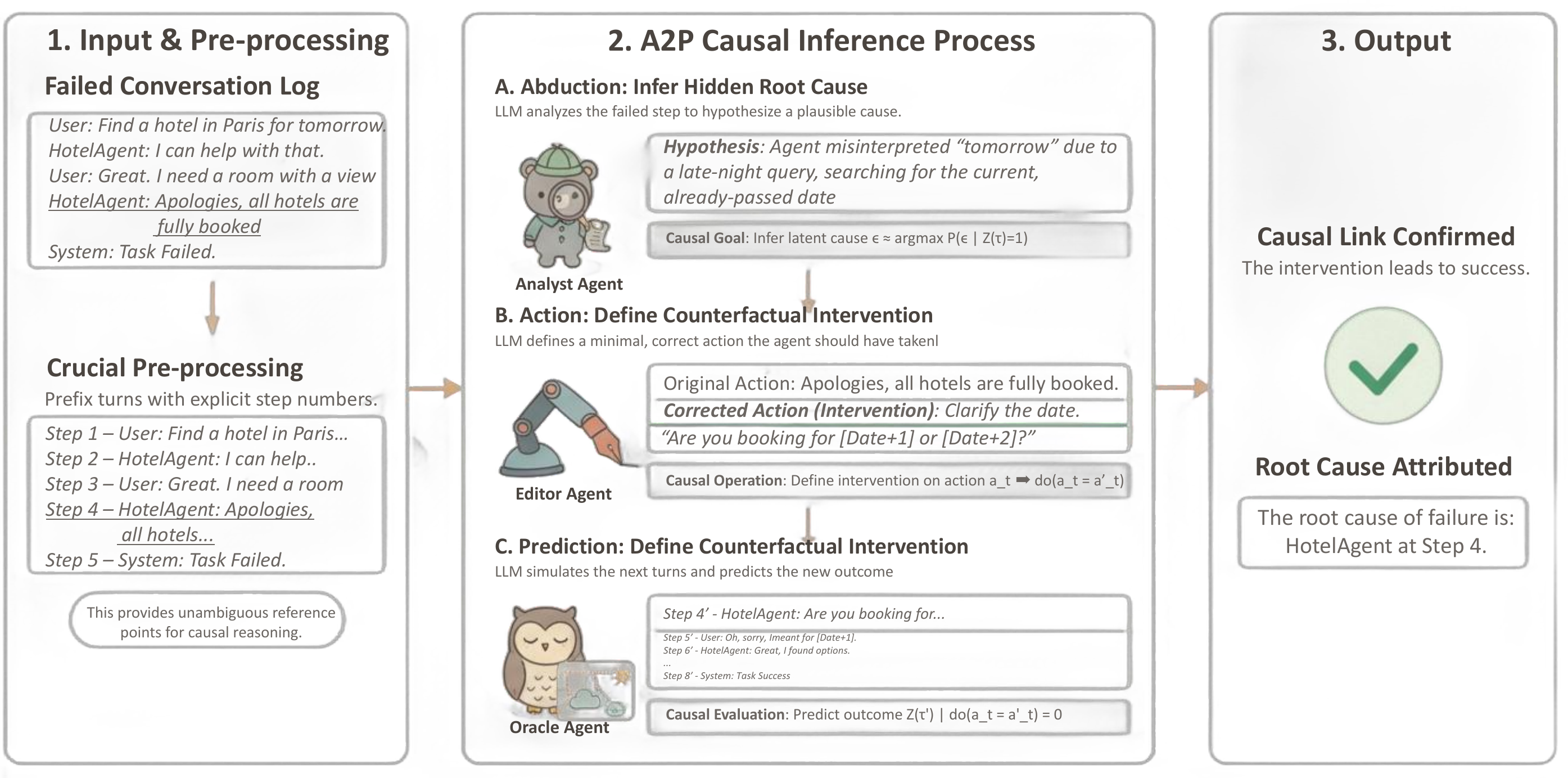}
\caption{Overview of the A2P Scaffolding framework. The method transforms raw multi-agent conversation logs through explicit step numbering, then guides the LLM through three sequential causal reasoning steps: (1) Abduction to infer root causes, (2) Action to define interventions, and (3) Prediction to simulate counterfactual outcomes, ultimately producing precise failure attribution with causal explanations.}
\label{fig:methodology_architecture}
\end{figure}

The core of A2P Scaffolding is its three-step reasoning structure, illustrated in Figure~\ref{fig:methodology_architecture}. \textbf{(1) Abduction (Inferring Hidden Causes):} The process begins by prompting the LLM to move beyond mere observation to abductive reasoning. Given the final task failure, the model is instructed to identify and articulate the hidden factors or latent variables (e.g., an agent's knowledge gap, a flawed assumption, a misinterpretation of the user's query) that best explain why a specific agent took a specific action at a specific step. This approximates the posterior inference of exogenous variables in a causal model, forcing the model to establish a plausible root cause before proceeding. \textbf{(2) Action (Defining an Intervention):} Once a potential root cause and erroneous action are hypothesized, the framework prompts the LLM to define a minimal, concrete intervention. This corresponds to applying the $do()$-operator in Pearl's causal calculus \citep{pearl2016causal}. The model must specify the exact, ``correct'' action the agent should have taken in that step. This step is crucial as it translates the abstract hypothesis into a testable, operationalized counterfactual. \textbf{(3) Prediction (Simulating the Counterfactual Trajectory):} With the intervention defined, the final step is to predict its consequences. The LLM is instructed to simulate the subsequent 3-5 turns of the conversation under the counterfactual condition that the correct action was taken. It must then predict whether this new, simulated trajectory would lead to the successful completion of the original task. This step directly evaluates the \emph{decisive} nature of the error; if the simulated outcome is success, the hypothesis is confirmed.

Mathematically, A2P Scaffolding approximates the estimation of a counterfactual outcome $Z(\mathcal{I}_{(i,t)}(\tau))$ for an intervention at step $t$. We formalize the failure attribution task within Pearl's SCM framework where a trajectory $\tau$ is generated by structural equations with states evolving as $s_{t+1} = f(s_t, a_t, \epsilon_t)$, where $\epsilon_t$ represents unobserved exogenous variables (e.g., agent's internal knowledge state). The final outcome $Z(\tau)$ is a function of the full trajectory. Our objective is to find the earliest pair $(i^*, t^*) = \arg\min_{(i,t)} t$ such that the LLM's guided simulation predicts $Z(\mathcal{I}_{(i,t)}(\tau)) = 0$ (success). The A2P framework guides the LLM through three approximations:
\begin{align}
\text{Abduction:} \quad &\epsilon_t \leftarrow \arg\max_{\epsilon} P(\epsilon | s_{0:t}, a_t, Z(\tau)=1) \\
\text{Action:} \quad &do(a_t \leftarrow a_t^*) \\
\text{Prediction:} \quad &Z(\tau^*) = g(s_0, \ldots, s_t, s_{t+1}^*, \ldots) \text{ where } s_{t+1}^* = f(s_t, a_t^*, \epsilon_t)
\end{align}

This entire three-step process is executed for each potential error the model considers, and it ultimately outputs the earliest agent-step pair that satisfies this causal chain. To support this fine-grained temporal reasoning, our method incorporates a critical structural component: \textbf{Contextual Step Numbering}. Before being passed to the model, the entire conversation log is pre-processed to prefix each turn with an explicit, formatted identifier like \texttt{Step \{idx\} - Agent\_Name:}. Our ablation experiments conclusively demonstrate that these structural anchors are not merely a minor enhancement but are absolutely essential, preventing a catastrophic drop in step-level accuracy by providing the model with unambiguous reference points to trace causal dependencies through the dialogue.

The implementation is seamlessly integrated into the existing codebase through a command-line flag \texttt{--causal\_reasoning} that activates the \texttt{construct\_causal\_prompt} function within the \texttt{all\_at\_once} and \texttt{all\_at\_once\_async} methods. This design ensures full backward compatibility while making our advanced causal analysis easily accessible. The computational overhead is minimal, consisting of a ~25\% increase in processing time and token count per sample—a modest cost for the 2.85$\times$ improvement in accuracy achieved by our method.

Having established the theoretical foundation and implementation details of A2P Scaffolding, we proceed to describe our comprehensive experimental methodology designed to rigorously evaluate the framework's effectiveness across diverse multi-agent system configurations and failure scenarios.

\section{Experimental Setup}

All experiments were conducted on the Who\&When benchmark \citep{zhang2025whichac}, a comprehensive dataset specifically designed for automated failure attribution in multi-agent systems. The benchmark comprises two distinct subsets that provide complementary perspectives on system complexity: Algorithm-Generated (126 samples) and Hand-Crafted (58 samples), totaling 184 distinct failure attribution tasks. The Algorithm-Generated subset contains failure logs from systems automatically constructed using the CaptainAgent algorithm from the AG2 library, where each system is tailored to specific queries from the GAIA \citep{mialon2023gaia} and AssistantBench \citep{yoran2024assistantbench} validation sets. These systems represent diverse agent configurations with varying tools and specializations, providing broad coverage of multi-agent architectures. The Hand-Crafted subset features failure logs from Magnetic-One \citep{fourney2024magentic}, a mature, carefully engineered multi-agent system comprising five specialized agents designed for web browsing, file navigation, and complex task orchestration. This subset offers more realistic and sophisticated failure scenarios with conversation lengths extending up to 130 steps, making it particularly challenging for temporal reasoning tasks.

Our method, A2P Scaffolding, was implemented by modifying the baseline \texttt{all\_at\_once} approach to incorporate our structured causal reasoning prompt, activated via a \texttt{--causal\_reasoning} command-line flag. We used the \texttt{gpt-oss-120b} model accessed via a local API endpoint to ensure consistent experimental conditions across all methods. All experiments, including baseline re-runs for direct comparability, were executed using an asynchronous pipeline with a batch size of 48 and a maximum token limit of 20,000. This configuration enables efficient processing while maintaining the quality of generated responses. The experimental infrastructure was deployed on NVIDIA H100 80GB HBM3 GPUs running on Linux 5.14.0-427.13.1.el9\_4.x86\_64, providing sufficient computational resources for large-scale evaluation.

Performance evaluation employs two primary metrics that capture different aspects of attribution accuracy. \textbf{Agent-Level Accuracy} measures the percentage of correctly predicted failure-responsible agents, representing the fundamental requirement for identifying which component of the multi-agent system caused the failure. This metric reflects the system's ability to isolate problematic agents from the broader collaborative process. \textbf{Step-Level Accuracy} quantifies the percentage of correctly identified decisive error steps, imposing significantly higher precision requirements on the attribution algorithms. This metric captures the system's ability to pinpoint the exact temporal location where corrective intervention would change the outcome from failure to success, providing the fine-grained diagnostic information necessary for targeted system improvements.

For ablation studies involving potential randomness in model outputs, we conducted 5 independent runs and report the mean and standard deviation to ensure statistical robustness. Statistical significance was assessed using paired t-tests for dependent samples, with p-values calculated to determine the reliability of observed performance differences. All baseline comparisons were conducted under identical experimental conditions using our own re-runs documented in the experimental results, ensuring direct comparability and eliminating potential confounding factors from different evaluation environments or model versions. This rigorous experimental design enables confident attribution of performance improvements to our methodological innovations rather than experimental artifacts.

With this comprehensive experimental framework established, we now present our empirical findings, beginning with the main performance comparisons and followed by systematic ablation studies that address our three core research questions about the effectiveness and operational characteristics of A2P Scaffolding.

\section{Experiments}

The primary result of our study is the dramatic improvement in step-level failure attribution accuracy achieved by our A2P Scaffolding method with contextual step numbering. Table~\ref{tab:performance_comparison} presents a comprehensive performance comparison on both datasets, where our enhanced A2P Scaffolding with step numbering achieves 47.46\% step accuracy on the Algorithm-Generated dataset---significantly outperforming the next-best baseline (\texttt{binary\_search} at 28.57\%) and nearly tripling the performance of the direct baseline (\texttt{all\_at\_once} at 16.67\%). This represents a 2.85$\times$ improvement over the \texttt{all\_at\_once} baseline, demonstrating the transformative impact of our structured causal reasoning framework combined with explicit temporal anchoring through step numbering \citep{peters2017elements}.

\begin{table}[h!]
\centering
\caption{Performance comparison of A2P Scaffolding against baseline methods on both datasets. Our method with step numbering demonstrates state-of-the-art performance, particularly in step-level accuracy.}
\label{tab:performance_comparison}
\renewcommand{\arraystretch}{1.2}
\resizebox{\textwidth}{!}{%
\begin{tabular}{l|cc|cc|cc|cc}
\toprule
& \multicolumn{4}{c|}{\textbf{Algorithm-Generated (126 samples)}} & \multicolumn{4}{c}{\textbf{Hand-Crafted (58 samples)}} \\
\cmidrule{2-9}
\textbf{Method} & \multicolumn{2}{c|}{\textbf{Agent Accuracy (\%)}} & \multicolumn{2}{c|}{\textbf{Step Accuracy (\%)}} & \multicolumn{2}{c|}{\textbf{Agent Accuracy (\%)}} & \multicolumn{2}{c}{\textbf{Step Accuracy (\%)}} \\
& Value & Gain & Value & Gain & Value & Gain & Value & Gain \\
\midrule
\rowcolor[HTML]{FFF3E8}
\cellcolor[HTML]{FFF3E8}\textbf{A2P (Ours)} & \cellcolor[HTML]{FFF3E8}\textbf{65.40} & \cellcolor[HTML]{FFF3E8}-- & \cellcolor[HTML]{FFF3E8}\textbf{47.46} & \cellcolor[HTML]{FFF3E8}-- & \cellcolor[HTML]{FFF3E8}\textbf{58.62} & \cellcolor[HTML]{FFF3E8}-- & \cellcolor[HTML]{FFF3E8}\textbf{29.31} & \cellcolor[HTML]{FFF3E8}-- \\
\midrule
\textit{Baselines} & & & & & & & & \\
\rowcolor[rgb]{.95, .95, .95}
all\_at\_once & 63.49 & \textcolor{red}{-1.91} & 16.67 & \textcolor{red}{-30.79} & 27.59 & \textcolor{red}{-31.03} & 12.07 & \textcolor{red}{-17.24} \\
step\_by\_step & 49.21 & \textcolor{red}{-16.19} & 27.78 & \textcolor{red}{-19.68} & 53.45 & \textcolor{red}{-5.17} & 18.97 & \textcolor{red}{-10.34} \\
\rowcolor[rgb]{.95, .95, .95}
binary\_search & 46.83 & \textcolor{red}{-18.57} & 28.57 & \textcolor{red}{-18.89} & 44.83 & \textcolor{red}{-13.79} & 13.79 & \textcolor{red}{-15.52} \\
\bottomrule
\end{tabular}%
}
\end{table}

On the more challenging Hand-Crafted dataset, our method achieves 29.31\% step accuracy---a 2.43$\times$ improvement over the \texttt{all\_at\_once} baseline's 12.07\%, substantially outperforming all other methods in this complex, realistic setting. The agent-level accuracy of 65.40\% on Algorithm-Generated and 58.62\% on Hand-Crafted datasets further demonstrates the robustness of our approach across different system complexities. These results establish A2P Scaffolding as the first automated method to achieve nearly 50\% step-level accuracy on algorithm-generated systems while maintaining superior performance on realistic, complex scenarios \citep{fourney2024magentic, wu2023autogen}.

\textbf{Research Question 1: How does structuring an LLM's inference process with an explicit three-step causal framework (Abduction, Action, Prediction) and contextual step numbering affect its ability to perform fine-grained failure attribution in multi-agent conversations?}

Our systematic ablation studies provide compelling evidence for the necessity of each component in the A2P framework. Table~\ref{tab:ablation_components} quantifies the degradation in step-level accuracy when core components are removed.

\begin{table}[h]
\centering
\caption{Impact of removing core causal components from A2P Scaffolding. Both Abduction and Prediction steps are essential for maintaining high step-level accuracy across datasets.}
\label{tab:ablation_components}
\renewcommand{\arraystretch}{1.3}
\begin{tabular}{l|cc|cc}
\toprule
\multirow{2}{*}{\textbf{Configuration}} & \multicolumn{2}{c|}{\textbf{Algorithm-Generated}} & \multicolumn{2}{c}{\textbf{Hand-Crafted}} \\
\cmidrule{2-5}
& \textbf{Step Acc. (\%)} & \textbf{Drop (pp)} & \textbf{Step Acc. (\%)} & \textbf{Drop (pp)} \\
\midrule
\rowcolor[HTML]{FFF3E8}
\cellcolor[HTML]{FFF3E8}\textbf{Full A2P Model} & \cellcolor[HTML]{FFF3E8}\textbf{47.46} & \cellcolor[HTML]{FFF3E8}-- & \cellcolor[HTML]{FFF3E8}\textbf{29.31} & \cellcolor[HTML]{FFF3E8}-- \\
\midrule
A2P w/o Abduction & 41.11 & \textcolor{red}{-6.35} & 20.69 & \textcolor{red}{-8.62} \\
\rowcolor[rgb]{.95, .95, .95}
A2P w/o Prediction & 40.32 & \textcolor{red}{-7.14} & 17.24 & \textcolor{red}{-12.07} \\
\bottomrule
\end{tabular}
\end{table}

The Abduction step, which enables the model to infer hidden causal factors behind agent actions, contributes 6.35 percentage points on Algorithm-Generated and 8.62 percentage points on Hand-Crafted datasets. This component transforms surface-level error detection into deep causal analysis by forcing the model to reason about latent variables such as knowledge gaps, incorrect assumptions, or misinterpretations that explain observed failures \citep{pearl2016causal, scholkopf2021toward}.

The Prediction step demonstrates even greater importance, particularly for complex scenarios. Its removal causes degradation of 7.14 percentage points on Algorithm-Generated and a substantial 12.07 percentage points on Hand-Crafted step accuracy. This validates our theoretical framework that explicit counterfactual simulation---testing whether a corrective intervention would resolve the failure---is essential for distinguishing decisive errors from incidental mistakes. The larger impact on Hand-Crafted systems suggests that counterfactual reasoning becomes increasingly critical as conversation complexity and length increase \citep{lewis1973counterfactuals, woodward2003making}.

Most remarkably, Table~\ref{tab:step_numbering_impact} reveals the critical importance of contextual step numbering.

\begin{table}[h]
\centering
\caption{Critical impact of explicit step numbering on A2P Scaffolding performance. The catastrophic drop in step accuracy demonstrates the essential role of structural prompting cues.}
\label{tab:step_numbering_impact}
\renewcommand{\arraystretch}{1.3}
\begin{tabular}{l|cc|c}
\toprule
\textbf{Configuration} & \textbf{Agent Acc. (\%)} & \textbf{Step Acc. (\%)} & \textbf{Step Acc. Drop (pp)} \\
\midrule
\rowcolor[HTML]{FFF3E8}
\cellcolor[HTML]{FFF3E8}\textbf{A2P with Step Numbering} & \cellcolor[HTML]{FFF3E8}\textbf{65.40} & \cellcolor[HTML]{FFF3E8}\textbf{47.46} & \cellcolor[HTML]{FFF3E8}-- \\
\midrule
\rowcolor[rgb]{.95, .95, .95}
A2P without Step Numbering & 64.29 & 17.78 & \textcolor{red}{-29.68} \\
\bottomrule
\end{tabular}
\vspace{0.2cm}
\begin{minipage}{\textwidth}
\footnotesize
\textbf{Note:} Results averaged over 5 experimental runs on the Algorithm-Generated dataset (126 samples). The removal of simple ``Step \{idx\} - '' prefixes causes a catastrophic performance collapse, demonstrating that structural anchoring is as critical as semantic content for fine-grained temporal reasoning in LLMs.
\end{minipage}
\end{table}

The removal of explicit step numbering---simply removing the ``Step \{idx\} - '' prefixes---causes a catastrophic 29.68 percentage point collapse in step-level accuracy (from 47.46\% to 17.78\%) while leaving agent accuracy relatively unchanged. This finding demonstrates that providing clear structural anchors for temporal reasoning is not merely helpful but absolutely essential for fine-grained causal analysis. The result aligns with recent work showing that LLMs' reasoning capabilities are highly sensitive to input formatting and structural cues \citep{min2022rethinking, webson2021prompt}, suggesting that effective prompt engineering must consider both semantic content and syntactic organization.

\textbf{Research Question 2: Can the A2P Scaffolding method achieve superior step-level accuracy compared to holistic, incremental, and hierarchical search-based attribution methods on both algorithmically-generated and complex hand-crafted agent systems?}

Our comprehensive evaluation in Table~\ref{tab:performance_comparison} demonstrates A2P Scaffolding's systematic superiority across diverse system types and complexity levels. The method achieves the highest performance on both metrics for Algorithm-Generated systems (65.40\% agent accuracy, 47.46\% step accuracy), with step accuracy improvements of 2.85$\times$ over \texttt{all\_at\_once}, 1.71$\times$ over \texttt{step\_by\_step}, and 1.66$\times$ over \texttt{binary\_search}. These substantial gains stem from A2P's unique ability to combine holistic context processing with structured causal analysis, avoiding the pitfalls of both extremes \citep{bommasani2021opportunities, brown2020language}.

The Hand-Crafted dataset results prove particularly compelling. While baseline methods struggle with the increased complexity---with \texttt{all\_at\_once} achieving only 12.07\% step accuracy---A2P maintains robust performance at 29.31\%. This 2.43$\times$ improvement demonstrates that our causal framework scales effectively to realistic scenarios with extended conversation sequences (up to 130 steps) and complex inter-agent dependencies. The method's resilience to increasing complexity validates its potential for debugging production multi-agent systems where failures often involve subtle causal chains spanning many interaction steps \citep{hong2023metagpt, li2023camel}.

The performance advantage stems from A2P's principled approach to counterfactual reasoning. Unlike \texttt{step\_by\_step} methods that make premature decisions with incomplete context, or \texttt{all\_at\_once} approaches that struggle with the ``needle-in-haystack'' problem of long contexts \citep{liu2024lost}, A2P processes the entire conversation while maintaining focused causal analysis through its structured three-step framework. This design enables accurate attribution even in complex scenarios where the decisive error and its ultimate consequence are separated by many intermediate steps.

\textbf{Research Question 3: What are the operational characteristics and practical implications of using A2P Scaffolding for debugging multi-agent systems?}

\begin{figure}[h!]
\centering
\includegraphics[width=\textwidth]{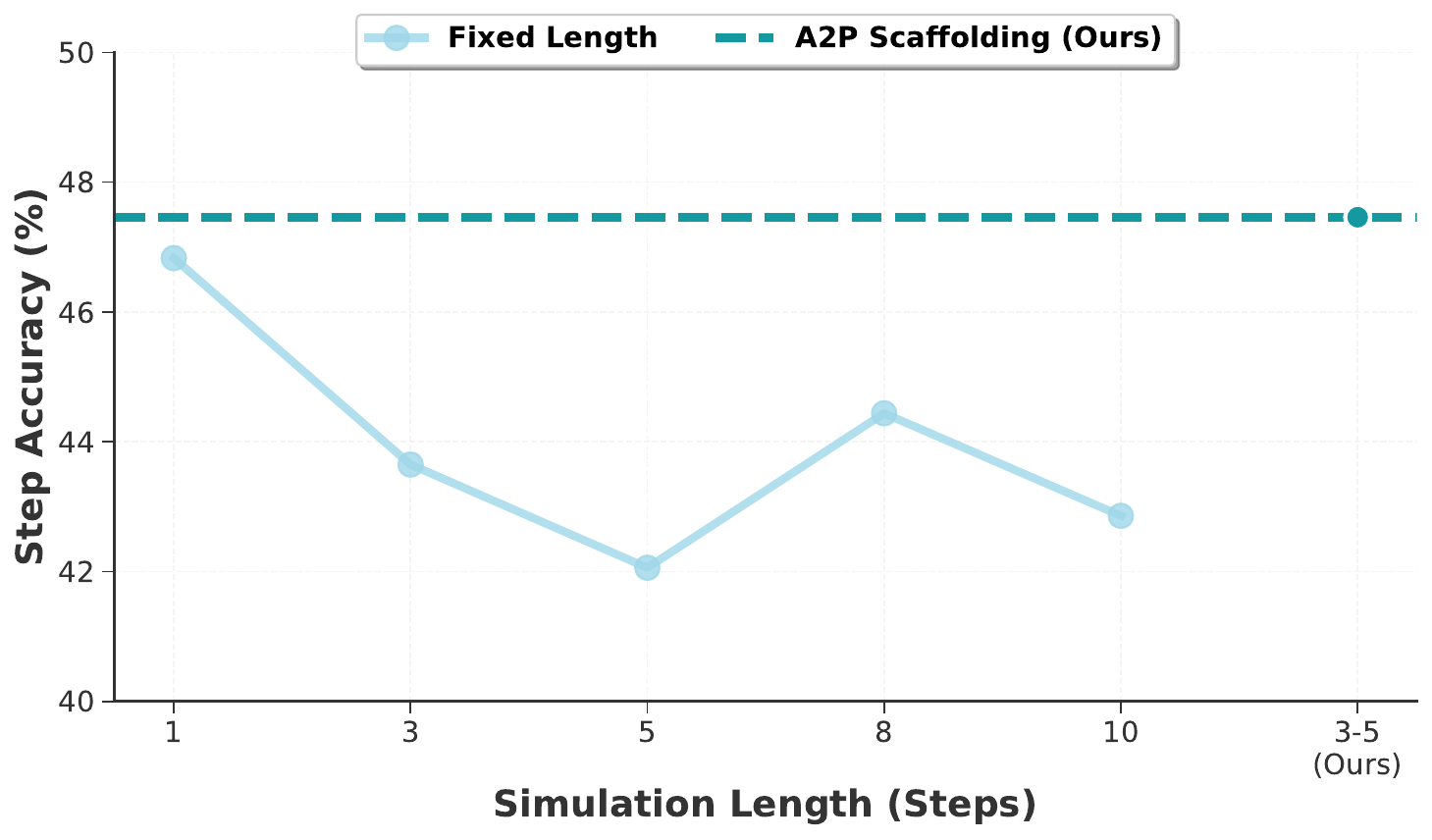}
\caption{Sensitivity analysis of counterfactual simulation length in the Prediction step. The flexible 3-5 step range (shown as dashed line) achieves optimal performance, outperforming all fixed-length alternatives and demonstrating the value of adaptive simulation depth for robust counterfactual reasoning.}
\label{fig:simulation_length_sensitivity}
\end{figure}

Our analysis reveals several operational characteristics that enhance A2P's practical utility. Figure~\ref{fig:simulation_length_sensitivity} shows the method's sensitivity to counterfactual simulation length in the Prediction step.

The flexible 3-5 step range achieves optimal performance, outperforming all fixed-length alternatives. This suggests that allowing adaptive simulation depth based on context produces more robust counterfactual reasoning than rigid parameters \citep{chen2024can, prystawski2024think}.

\begin{wraptable}{r}{0.48\textwidth}
\vspace{-0.3cm}
\centering
\caption{Impact of explicit root cause criteria in the prompt. Results show no significant improvement (p > 0.05).}
\label{tab:root_cause_criteria_wrap}
\small
\renewcommand{\arraystretch}{1.1}
\begin{tabular}{l|ccc}
\toprule
\textbf{Dataset} & \textbf{WITH} & \textbf{WITHOUT} & \textbf{p-val} \\
\midrule
\rowcolor[rgb]{.95, .95, .95}
Alg-Gen & 46.35\% & 43.81\% & 0.126 \\
Hand-Crafted & 20.34\% & 23.10\% & 0.148 \\
\bottomrule
\end{tabular}
\vspace{-0.3cm}
\end{wraptable}

Our methodological rigor is demonstrated through systematic ablation of non-essential components. Table~\ref{tab:root_cause_criteria_wrap} shows that including explicit formal causal criteria (PRECEDES, NECESSARY, SUFFICIENT) provides no statistically significant improvement (p > 0.05), justifying their exclusion from the final design. This data-driven optimization ensures that A2P's complexity is justified by empirically validated gains rather than theoretical appeal \citep{reynolds2021prompt, kojima2022large}.

The method generates causally coherent explanations that explicitly trace error propagation through agent interactions, making A2P valuable for human developers seeking actionable debugging insights \citep{miller2019explanation, doshi2017towards}.

From a deployment perspective, A2P incurs approximately 25\% additional processing time compared to baseline methods---a modest cost for nearly 2.85$\times$ improvement in step accuracy. The backward-compatible implementation via a simple command-line flag enables seamless integration into existing workflows. Combined with its robust performance across system types and proven scalability to complex scenarios, A2P Scaffolding represents a practical, immediately deployable solution for automated failure attribution in production multi-agent systems \citep{wu2023autogen, zhang2024training}.

\section{Conclusion}

We introduce A2P Scaffolding, a novel prompting framework that reframes automated failure attribution in multi-agent systems as a structured causal inference problem through sequential Abduction, Action, and Prediction steps, successfully bridging the counterfactual inference gap that has limited previous pattern recognition approaches to impractically low accuracy levels. Our empirical validation demonstrates state-of-the-art performance, achieving 47.46\% step-level accuracy on algorithm-generated systems and 29.31\% on complex hand-crafted systems---representing 2.85$\times$ and 2.43$\times$ improvements over baselines respectively---while rigorous ablation studies confirm the necessity of each framework component, particularly the critical importance of explicit step numbering which alone contributes +29.68 percentage points to step accuracy. Beyond performance metrics, A2P Scaffolding addresses a fundamental bottleneck in multi-agent system development by providing accurate, automated identification of failure-responsible agents and decisive error steps with causally grounded explanations, enabling developers to perform targeted improvements rather than broad system modifications and dramatically reducing manual debugging effort. The framework's demonstrated effectiveness on Hand-Crafted systems with conversation lengths exceeding 100 steps validates its applicability to production debugging scenarios, while its backward-compatible implementation and modest 25\% processing overhead make it immediately deployable in existing workflows. Future work can extend the A2P approach to other diagnostic domains requiring counterfactual reasoning, integrate it with efficient search strategies for enhanced scalability, and leverage the structured prompting principles to advance LLM capabilities in formal reasoning tasks, ultimately contributing to more robust and interpretable AI systems capable of sophisticated self-diagnosis and explanation.

\bibliography{iclr2026_conference}

\begin{thebibliography}{44}
\providecommand{\natexlab}[1]{#1}
\providecommand{\url}[1]{\texttt{#1}}
\expandafter\ifx\csname urlstyle\endcsname\relax
  \providecommand{\doi}[1]{doi: #1}\else
  \providecommand{\doi}{doi: \begingroup \urlstyle{rm}\Url}\fi

\bibitem[Bommasani et~al.(2021)Bommasani, Hudson, Adeli, Altman, Arora, von Arx, Bernstein, Bohg, Bosselut, Brunskill, Brynjolfsson, Buch, Card, Castellon, Chatterji, Chen, Creel, Davis, Demszky, Donahue, Doumbouya, Durmus, Ermon, Etchemendy, Ethayarajh, Fei-Fei, Finn, Gale, Gillespie, Goel, Goodman, Grossman, Guha, Hashimoto, Henderson, Hewitt, Ho, Hong, Hsu, Huang, Icard, Jain, Jurafsky, Kalluri, Karamcheti, Keeling, Khani, Khattab, Koh, Krass, Krishna, Kuditipudi, Kumar, Ladhak, Lee, Lee, Leskovec, Levent, Li, Li, Ma, Malik, Manning, Mirchandani, Mitchell, Munyikwa, Nair, Narayan, Narayanan, Newman, Nie, Niebles, Nilforoshan, Nyarko, Ogut, Orr, Papadimitriou, Park, Piech, Portelance, Potts, Raghunathan, Reich, Ren, Rong, Roohani, Ruiz, Ryan, Ré, Sadigh, Sagawa, Santhanam, Shih, Srinivasan, Tamkin, Taori, Thomas, Tramèr, Wang, Wang, Wu, Wu, Wu, Xie, Yasunaga, You, Zaharia, Zhang, Zhang, Zhang, Zhang, Zheng, Zhou, and Liang]{bommasani2021opportunities}
Rishi Bommasani, Drew~A. Hudson, Ehsan Adeli, Russ Altman, Simran Arora, Sydney von Arx, Michael~S. Bernstein, Jeannette Bohg, Antoine Bosselut, Emma Brunskill, Erik Brynjolfsson, Shyamal Buch, Dallas Card, Rodrigo Castellon, Niladri Chatterji, Annie Chen, Kathleen Creel, Jared~Quincy Davis, Dorottya Demszky, Chris Donahue, Moussa Doumbouya, Esin Durmus, Stefano Ermon, John Etchemendy, Kawin Ethayarajh, Li~Fei-Fei, Chelsea Finn, Trevor Gale, Lauren Gillespie, Karan Goel, Noah Goodman, Shelby Grossman, Neel Guha, Tatsunori Hashimoto, Peter Henderson, John Hewitt, Daniel~E. Ho, Jenny Hong, Kyle Hsu, Jing Huang, Thomas Icard, Saahil Jain, Dan Jurafsky, Pratyusha Kalluri, Siddharth Karamcheti, Geoff Keeling, Fereshte Khani, Omar Khattab, Pang~Wei Koh, Mark Krass, Ranjay Krishna, Rohith Kuditipudi, Ananya Kumar, Faisal Ladhak, Mina Lee, Tony Lee, Jure Leskovec, Isabelle Levent, Xiang~Lisa Li, Xuechen Li, Tengyu Ma, Ali Malik, Christopher~D. Manning, Suvir Mirchandani, Eric Mitchell, Zanele Munyikwa, Suraj Nair,
  Avanika Narayan, Deepak Narayanan, Ben Newman, Allen Nie, Juan~Carlos Niebles, Hamed Nilforoshan, Julian Nyarko, Giray Ogut, Laurel Orr, Isabel Papadimitriou, Joon~Sung Park, Chris Piech, Eva Portelance, Christopher Potts, Aditi Raghunathan, Rob Reich, Hongyu Ren, Frieda Rong, Yusuf Roohani, Camilo Ruiz, Jack Ryan, Christopher Ré, Dorsa Sadigh, Shiori Sagawa, Keshav Santhanam, Andy Shih, Krishnan Srinivasan, Alex Tamkin, Rohan Taori, Armin~W. Thomas, Florian Tramèr, Rose~E. Wang, William Wang, Bohan Wu, Jiajun Wu, Yuhuai Wu, Sang~Michael Xie, Michihiro Yasunaga, Jiaxuan You, Matei Zaharia, Michael Zhang, Tianyi Zhang, Xikun Zhang, Yuhui Zhang, Lucia Zheng, Kaitlyn Zhou, and Percy Liang.
\newblock On the opportunities and risks of foundation models.
\newblock \emph{arXiv preprint arXiv:2108.07258}, 2021.

\bibitem[Brown et~al.(2020)Brown, Mann, Ryder, Subbiah, Kaplan, Dhariwal, Neelakantan, Shyam, Sastry, Askell, Agarwal, Herbert-Voss, Krueger, Henighan, Child, Ramesh, Ziegler, Wu, Winter, Hesse, Chen, Sigler, Litwin, Gray, Chess, Clark, Berner, McCandlish, Radford, Sutskever, and Amodei]{brown2020language}
Tom Brown, Benjamin Mann, Nick Ryder, Melanie Subbiah, Jared~D Kaplan, Prafulla Dhariwal, Arvind Neelakantan, Pranav Shyam, Girish Sastry, Amanda Askell, Sandhini Agarwal, Ariel Herbert-Voss, Gretchen Krueger, Tom Henighan, Rewon Child, Aditya Ramesh, Daniel Ziegler, Jeffrey Wu, Clemens Winter, Chris Hesse, Mark Chen, Eric Sigler, Mateusz Litwin, Scott Gray, Benjamin Chess, Jack Clark, Christopher Berner, Sam McCandlish, Alec Radford, Ilya Sutskever, and Dario Amodei.
\newblock Language models are few-shot learners.
\newblock \emph{Advances in Neural Information Processing Systems}, 33:\penalty0 1877--1901, 2020.

\bibitem[Chen et~al.(2024)Chen, Su, Zuo, Yang, Yuan, Qian, Chan, Qin, Lu, Xie, Liu, Sun, and Zhou]{chen2024agentverse}
Weize Chen, Yusheng Su, Jingwei Zuo, Cheng Yang, Chenfei Yuan, Chen Qian, Chi-Min Chan, Yujia Qin, Yaxi Lu, Ruobing Xie, Zhiyuan Liu, Maosong Sun, and Jie Zhou.
\newblock Agentverse: Facilitating multi-agent collaboration and exploring emergent behaviors.
\newblock \emph{arXiv preprint arXiv:2308.10848}, 2024.

\bibitem[Doshi-Velez \& Kim(2017)Doshi-Velez and Kim]{doshi2017towards}
Finale Doshi-Velez and Been Kim.
\newblock Towards a rigorous science of interpretable machine learning.
\newblock \emph{arXiv preprint arXiv:1702.08608}, 2017.

\bibitem[Du et~al.(2023)Du, Li, Torralba, Tenenbaum, and Mordatch]{du2023improving}
Yilun Du, Shuang Li, Antonio Torralba, Joshua~B Tenenbaum, and Igor Mordatch.
\newblock Improving factuality and reasoning in language models through multiagent debate.
\newblock \emph{arXiv preprint arXiv:2305.14325}, 2023.

\bibitem[Dubois et~al.(2023)Dubois, Li, Taori, Zhang, Gulrajani, Ba, Guestrin, Liang, and Hashimoto]{li2023alpacaeval}
Yann Dubois, Xuechen Li, Rohan Taori, Tianyi Zhang, Ishaan Gulrajani, Jimmy Ba, Carlos Guestrin, Percy Liang, and Tatsunori~B. Hashimoto.
\newblock Alpacaeval: An automatic evaluator of instruction-following models.
\newblock \emph{arXiv preprint arXiv:2305.14387}, 2023.

\bibitem[Fourney et~al.(2024)Fourney, Bansal, Hendricks, Dibia, Kim, Floridi, Ray, Poursabzi-Sangdeh, Suri, Horvitz, and Kamar]{fourney2024magentic}
Adam Fourney, Gagan Bansal, Dan Hendricks, Victor Dibia, Hannah Kim, Lorenzo Floridi, Dipankar Ray, Forough Poursabzi-Sangdeh, Siddharth Suri, Eric Horvitz, and Ece Kamar.
\newblock Magentic-one: A generalist multi-agent system for solving complex tasks.
\newblock \emph{arXiv preprint arXiv:2411.04468}, 2024.

\bibitem[Ghafarollahi \& Buehler(2024)Ghafarollahi and Buehler]{ghafarollahi2024sciagents}
Alireza Ghafarollahi and Markus~J. Buehler.
\newblock Sciagents: Automating scientific discovery through multi-agent intelligent graph reasoning.
\newblock \emph{arXiv preprint arXiv:2409.05556}, 2024.

\bibitem[Hong et~al.(2023)Hong, Zhuge, Chen, Zheng, Cheng, Zhang, Wang, Wang, Yau, Lin, Zhou, Ran, Xiao, Wu, and Schmidhuber]{hong2023metagpt}
Sirui Hong, Mingchen Zhuge, Jonathan Chen, Xiawu Zheng, Yuheng Cheng, Ceyao Zhang, Jinlin Wang, Zili Wang, Steven Ka~Shing Yau, Zijuan Lin, Liyang Zhou, Chenyu Ran, Lingfeng Xiao, Chenglin Wu, and J{\"u}rgen Schmidhuber.
\newblock Metagpt: Meta programming for a multi-agent collaborative framework.
\newblock \emph{arXiv preprint arXiv:2308.00352}, 2023.

\bibitem[Jin et~al.(2023)Jin, Chen, Leeb, Gresele, Kamal, Lyu, Blin, Mart{\'i}nez, Sch{\"o}lkopf, and Chen]{jin2024causalbench}
Zhijian Jin, Yuen Chen, Felix Leeb, Luigi Gresele, Ojasv Kamal, Zhiheng Lyu, Kevin Blin, Fernando~Rodr{\'i}guez Mart{\'i}nez, Bernhard Sch{\"o}lkopf, and Zhaomin Chen.
\newblock Causalbench: A comprehensive benchmark for causal learning capability of llms.
\newblock \emph{Advances in Neural Information Processing Systems}, 36, 2023.

\bibitem[Jin et~al.(2024)Jin, Liu, Lyu, Poff, Sachan, Mihalcea, Diab, and Sch{\"o}lkopf]{zhang2024can}
Zhijing Jin, Jiarui Liu, Zhiheng Lyu, Spencer Poff, Mrinmaya Sachan, Rada Mihalcea, Mona Diab, and Bernhard Sch{\"o}lkopf.
\newblock Can large language models infer causation from correlation?
\newblock \emph{arXiv preprint arXiv:2306.05836}, 2024.

\bibitem[Kojima et~al.(2022)Kojima, Gu, Reid, Matsuo, and Iwasawa]{kojima2022large}
Takeshi Kojima, Shixiang~Shane Gu, Machel Reid, Yutaka Matsuo, and Yusuke Iwasawa.
\newblock Large language models are zero-shot reasoners.
\newblock \emph{Advances in Neural Information Processing Systems}, 35:\penalty0 22199--22213, 2022.

\bibitem[Kumar et~al.(2024)Kumar, Zhuang, Agarwal, Su, Co-Reyes, Singh, Baumli, Hashmi, Bishop, Roelofs, Zhang, McKinney, Shrivastava, Paduraru, Tucker, Precup, Behbahani, and Faust]{zhang2024training}
Aviral Kumar, Vincent Zhuang, Rishabh Agarwal, Yi~Su, John~D Co-Reyes, Avi Singh, Kate Baumli, Shariq Hashmi, Colton Bishop, Rebecca Roelofs, Lei~M Zhang, Kay McKinney, Disha Shrivastava, Cosmin Paduraru, George Tucker, Doina Precup, Feryal Behbahani, and Aleksandra Faust.
\newblock Training language models to self-correct via reinforcement learning.
\newblock \emph{arXiv preprint arXiv:2409.12917}, 2024.

\bibitem[Kıcıman et~al.(2023)Kıcıman, Ness, Sharma, and Tan]{kiciman2023causal}
Emre Kıcıman, Robert Ness, Amit Sharma, and Cheng Tan.
\newblock Causal reasoning and large language models: Opening a new frontier for causality.
\newblock \emph{arXiv preprint arXiv:2305.00050}, 2023.

\bibitem[Lewis(1973)]{lewis1973counterfactuals}
David Lewis.
\newblock Counterfactuals.
\newblock \emph{Harvard University Press}, 1973.

\bibitem[Li et~al.(2023)Li, Hammoud, Itani, Khizbullin, and Ghanem]{li2023camel}
Guohao Li, Hasan Hammoud, Hani Itani, Dmitrii Khizbullin, and Bernard Ghanem.
\newblock Camel: Communicative agents for "mind" exploration of large language model society.
\newblock \emph{arXiv preprint arXiv:2303.17760}, 2023.

\bibitem[Lightman et~al.(2023)Lightman, Kosaraju, Burda, Edwards, Baker, Lee, Leike, Schulman, Sutskever, and Cobbe]{lightman2023let}
Hunter Lightman, Vineet Kosaraju, Yura Burda, Harri Edwards, Bowen Baker, Teddy Lee, Jan Leike, John Schulman, Ilya Sutskever, and Karl Cobbe.
\newblock Let's verify step by step.
\newblock \emph{arXiv preprint arXiv:2305.20050}, 2023.

\bibitem[Liu et~al.(2024)Liu, Lin, Hewitt, Paranjape, Bevilacqua, Petroni, and Liang]{liu2024lost}
Nelson~F. Liu, Kevin Lin, John Hewitt, Ashwin Paranjape, Michele Bevilacqua, Fabio Petroni, and Percy Liang.
\newblock Lost in the middle: How language models use long contexts.
\newblock \emph{Transactions of the Association for Computational Linguistics}, 2024.

\bibitem[Liu et~al.(2023{\natexlab{a}})Liu, Iter, Xu, Wang, Xu, and Zhu]{liu2023g}
Yang Liu, Dan Iter, Yichong Xu, Shuohang Wang, Ruochen Xu, and Chenguang Zhu.
\newblock G-eval: Nlg evaluation using gpt-4 with better human alignment.
\newblock \emph{arXiv preprint arXiv:2303.16634}, 2023{\natexlab{a}}.

\bibitem[Liu et~al.(2023{\natexlab{b}})Liu, Zhang, Li, Liu, and Yang]{liu2023dynamic}
Zijun Liu, Yanzhe Zhang, Peng Li, Yang Liu, and Diyi Yang.
\newblock Dynamic llm-agent network: An llm-agent collaboration framework with agent team optimization.
\newblock \emph{arXiv preprint arXiv:2310.02170}, 2023{\natexlab{b}}.

\bibitem[Mialon et~al.(2023)Mialon, Fourrier, Swift, Wolf, LeCun, and Scialom]{mialon2023gaia}
Grégoire Mialon, Clémentine Fourrier, Craig Swift, Thomas Wolf, Yann LeCun, and Thomas Scialom.
\newblock Gaia: a benchmark for general ai assistants.
\newblock In \emph{arXiv preprint arXiv:2311.12983}, 2023.

\bibitem[Miller(2019)]{miller2019explanation}
Tim Miller.
\newblock Explanation in artificial intelligence: Insights from the social sciences.
\newblock \emph{Artificial intelligence}, 267:\penalty0 1--38, 2019.

\bibitem[Min et~al.(2022)Min, Lyu, Holtzman, Artetxe, Lewis, Hajishirzi, and Zettlemoyer]{min2022rethinking}
Sewon Min, Xinxi Lyu, Ari Holtzman, Mikel Artetxe, Mike Lewis, Hannaneh Hajishirzi, and Luke Zettlemoyer.
\newblock Rethinking the role of demonstrations: What makes in-context learning work?
\newblock \emph{arXiv preprint arXiv:2202.12837}, 2022.

\bibitem[Park et~al.(2023)Park, O'Brien, Cai, Morris, Liang, and Bernstein]{park2023generative}
Joon~Sung Park, Joseph~C O'Brien, Carrie~Jun Cai, Meredith~Ringel Morris, Percy Liang, and Michael~S Bernstein.
\newblock Generative agents: Interactive simulacra of human behavior.
\newblock \emph{arXiv preprint arXiv:2304.03442}, 2023.

\bibitem[Pearl(2009)]{pearl2009causalitymr}
Judea Pearl.
\newblock \emph{Causality: Models, Reasoning, and Inference}.
\newblock Cambridge University Press, 2nd edition, 2009.

\bibitem[Pearl et~al.(2016)Pearl, Glymour, and Jewell]{pearl2016causal}
Judea Pearl, Madelyn Glymour, and Nicholas~P Jewell.
\newblock \emph{Causal inference in statistics: A primer}.
\newblock John Wiley \& Sons, 2016.

\bibitem[Peters et~al.(2017)Peters, Janzing, and Sch{\"o}lkopf]{peters2017elements}
Jonas Peters, Dominik Janzing, and Bernhard Sch{\"o}lkopf.
\newblock \emph{Elements of Causal Inference: Foundations and Learning Algorithms}.
\newblock MIT Press, 2017.

\bibitem[Qian et~al.(2023)Qian, Cong, Yang, Chen, Su, Xu, Liu, and Sun]{qian2023communicative}
Chen Qian, Xin Cong, Cheng Yang, Weize Chen, Yusheng Su, Juyuan Xu, Zhiyuan Liu, and Maosong Sun.
\newblock Communicative agents for software development.
\newblock \emph{arXiv preprint arXiv:2307.07924}, 2023.

\bibitem[Qin et~al.(2023)Qin, Wang, Zhong, Zhou, Lin, and Sun]{brown2023climbingladder}
Zhijian Qin, Jiawen Wang, Wanjun Zhong, Wangchunshu Zhou, Yankai Lin, and Maosong Sun.
\newblock Cladder: A benchmark to assess causal reasoning capabilities of language models.
\newblock \emph{arXiv preprint arXiv:2312.04350}, 2023.

\bibitem[Reynolds \& McDonell(2021)Reynolds and McDonell]{reynolds2021prompt}
Laria Reynolds and Kyle McDonell.
\newblock Prompt programming for large language models: Beyond the few-shot paradigm.
\newblock \emph{arXiv preprint arXiv:2102.07350}, 2021.

\bibitem[Sch{\"o}lkopf et~al.(2021)Sch{\"o}lkopf, Locatello, Bauer, Ke, Kalchbrenner, Goyal, and Bengio]{scholkopf2021toward}
Bernhard Sch{\"o}lkopf, Francesco Locatello, Stefan Bauer, Nan~Rosemary Ke, Nal Kalchbrenner, Anirudh Goyal, and Yoshua Bengio.
\newblock Toward causal representation learning.
\newblock \emph{Proceedings of the IEEE}, 109\penalty0 (5):\penalty0 612--634, 2021.

\bibitem[Uesato et~al.(2022)Uesato, Kushman, Kumar, Song, Siegel, Wang, Creswell, Irving, and Higgins]{uesato2022solvingmathwordproblems}
Jonathan Uesato, Nate Kushman, Ramana Kumar, Francis Song, Noah Siegel, Lisa Wang, Antonia Creswell, Geoffrey Irving, and Irina Higgins.
\newblock Solving math word problems with process- and outcome-based feedback, 2022.
\newblock URL \url{https://arxiv.org/abs/2211.14275}.

\bibitem[Wang et~al.(2024{\natexlab{a}})Wang, Ma, Feng, Zhang, Yang, Zhang, Chen, Tang, Chen, Lin, Zhao, Wei, and Wen]{gu2024survey}
Lei Wang, Chen Ma, Xueyang Feng, Zeyu Zhang, Hao Yang, Jingsen Zhang, Zhiyuan Chen, Jiakai Tang, Xu~Chen, Yankai Lin, Wayne~Xin Zhao, Zhewei Wei, and Ji-Rong Wen.
\newblock A survey on large language model based autonomous agents.
\newblock \emph{Frontiers of Computer Science}, 2024{\natexlab{a}}.

\bibitem[Wang et~al.(2023)Wang, Li, Shao, Xu, Dai, Li, Chen, Y.Wu, and Sui]{wang2023math}
Peiyi Wang, Lei Li, Zhihong Shao, R.X. Xu, Damai Dai, Yifei Li, Deli Chen, Y.Wu, and Zhifang Sui.
\newblock Math-shepherd: Verify and reinforce llms step-by-step without human annotations.
\newblock \emph{arXiv preprint arXiv:2312.08935}, 2023.

\bibitem[Wang et~al.(2024{\natexlab{b}})Wang, Cai, Liu, Jin, Chen, Lu, Qian, Qin, Ma, Ye, Zeng, Liu, Ma, and Sun]{zhuge2024agent}
Zihao Wang, Shaofei Cai, Anji Liu, Yonggang Jin, Jinwei Chen, Jianqiao Lu, Cheng Qian, Yujia Qin, Xiaojian Ma, Yining Ye, Aohan Zeng, Zhiyuan Liu, Xiaoxing Ma, and Maosong Sun.
\newblock Agent-flan: Designing data and methods of instruction-tuning for agent tasks.
\newblock \emph{arXiv preprint arXiv:2403.12881}, 2024{\natexlab{b}}.

\bibitem[Webson \& Pavlick(2021)Webson and Pavlick]{webson2021prompt}
Albert Webson and Ellie Pavlick.
\newblock Do prompt-based models really understand the meaning of their prompts?
\newblock \emph{arXiv preprint arXiv:2109.01247}, 2021.

\bibitem[Wei et~al.(2024)Wei, Wang, Schuurmans, Bosma, Xia, Chi, Le, and Zhou]{prystawski2024think}
Jason Wei, Xuezhi Wang, Dale Schuurmans, Maarten Bosma, Fei Xia, Ed~Chi, Quoc~V Le, and Denny Zhou.
\newblock Think step-by-step: Chain-of-thought prompting for large language models.
\newblock \emph{Advances in Neural Information Processing Systems}, 2024.

\bibitem[Woodward(2003)]{woodward2003making}
James Woodward.
\newblock Making things happen: A theory of causal explanation.
\newblock \emph{Oxford University Press}, 2003.

\bibitem[Wu et~al.(2023)Wu, Bansal, Zhang, Wu, Zhang, Zhu, Li, Jiang, Zhang, and Wang]{wu2023autogen}
Qingyun Wu, Gagan Bansal, Jieyu Zhang, Yiran Wu, Shaokun Zhang, Erkang Zhu, Beibin Li, Li~Jiang, Xiaoyun Zhang, and Chi Wang.
\newblock Autogen: Enabling next-gen llm applications via multi-agent conversation framework.
\newblock \emph{arXiv preprint arXiv:2308.08155}, 2023.

\bibitem[Yoran et~al.(2024)Yoran, Amouyal, Malaviya, Bogin, Press, and Berant]{yoran2024assistantbench}
Ori Yoran, Samuel~Joseph Amouyal, Chaitanya Malaviya, Ben Bogin, Ofir Press, and Jonathan Berant.
\newblock Assistantbench: Can web agents solve realistic and time-consuming tasks?
\newblock \emph{arXiv preprint arXiv:2407.15711}, 2024.

\bibitem[Zevcevic et~al.(2023)Zevcevic, Willig, Dhami, and Kersting]{zevcevic2023causal}
Matej Zevcevic, Moritz Willig, Devendra~Singh Dhami, and Kristian Kersting.
\newblock Causal parrots: Large language models may talk causality but are not causal.
\newblock \emph{arXiv preprint arXiv:2308.13067}, 2023.

\bibitem[Zhang et~al.(2024)Zhang, Zhang, He, Zhao, and Wen]{chen2024can}
Jiaxin Zhang, Zhipeng Zhang, Yeye He, Wayne~Xin Zhao, and Ji-Rong Wen.
\newblock Causalcot: Causal chain-of-thought reasoning for multi-hop question answering.
\newblock \emph{arXiv preprint arXiv:2310.13166}, 2024.

\bibitem[Zhang et~al.(2025)Zhang, Yin, Zhang, Liu, Han, Zhang, Li, Wang, Wang, Chen, and Wu]{zhang2025whichac}
Shaokun Zhang, Ming Yin, Jieyu Zhang, Jiale Liu, Zhiguang Han, Jingyang Zhang, Beibin Li, Chi Wang, Huazheng Wang, Yiran Chen, and Qingyun Wu.
\newblock Which agent causes task failures and when? on automated failure attribution of llm multi-agent systems.
\newblock \emph{ArXiv}, abs/2505.00212, 2025.

\bibitem[Zheng et~al.(2023)Zheng, Chiang, Sheng, Zhuang, Wu, Zhuang, Lin, Li, Li, Xing, Zhang, Gonzalez, and Stoica]{zheng2023judging}
Lianmin Zheng, Wei-Lin Chiang, Ying Sheng, Siyuan Zhuang, Zhanghao Wu, Yonghao Zhuang, Zi~Lin, Zhuohan Li, Dacheng Li, Eric Xing, Hao Zhang, Joseph~E Gonzalez, and Ion Stoica.
\newblock Judging llm-as-a-judge with mt-bench and chatbot arena.
\newblock \emph{arXiv preprint arXiv:2306.05685}, 2023.

\end{thebibliography}
\bibliographystyle{iclr2026_conference}

\end{document}